\documentclass{scrartcl}
\usepackage{geometry}
\PassOptionsToPackage{numbers, sort&compress}{natbib}
\usepackage{natbib}

\usepackage{times}
\usepackage{helvet}
\usepackage{courier}
\usepackage{multirow}

\usepackage[utf8]{inputenc} 
\usepackage[T1]{fontenc}    
\usepackage[colorlinks=true, citecolor=green]{hyperref}       
\usepackage{url}            
\usepackage{subcaption}

\usepackage{fullpage}
\usepackage{authblk}

\usepackage{longtable}
\usepackage{booktabs}
\usepackage{verbatim}
\usepackage{multirow}
\usepackage{lscape}
\usepackage{authblk}

\usepackage{tikz}
\usetikzlibrary{shapes,decorations}

\usepackage{amssymb}
\usepackage{nicefrac}       
\usepackage{amsfonts}       
\usepackage{amsmath}

\usepackage{mathtools}

\usepackage{siunitx}
\newcommand{\norm}[1]{\lVert#1\rVert}

\newcommand{\ip}[1]{\langle#1\rangle}

\DeclareMathOperator*{\E}{\mathbb{E}}

\DeclareMathOperator{\R}{\mathbb{R}}

\usepackage{mathtools}

\usetikzlibrary{positioning}

\usepackage{listings}
\usepackage{color}

\definecolor{dkgreen}{rgb}{0,0.6,0}
\definecolor{gray}{rgb}{0.5,0.5,0.5}
\definecolor{mauve}{rgb}{0.58,0,0.82}

\makeatletter
\def\set@curr@file#1{\def\@curr@file{#1}} 
\makeatother

\newcommand{\package}{\href{https://contextualized.ml}{ContextualizedML}}

\newcommand*\samethanks[1][\value{footnote}]{\footnotemark[#1]}

\usepackage{amsthm}

\begin{document}

\title{Contextualized Machine Learning}

\author[1,2]{Benjamin Lengerich\thanks{Equal contribution}\thanks{\texttt{blengeri@mit.edu}}}
\author[3]{Caleb N. Ellington\samethanks[1]\thanks{\texttt{cellingt@cs.cmu.edu}}}
\author[5]{Andrea Rubbi}
\author[1,2]{Manolis Kellis}
\author[3,4]{Eric P. Xing}

\affil[1]{Massachusetts Institute of Technology}
\affil[2]{Broad Institute of MIT and Harvard}
\affil[3]{Carnegie Mellon University}
\affil[4]{Mohamed Bin Zayed University of Artificial Intelligence}
\affil[5]{University of Cambridge}

\maketitle

\begin{abstract}
We examine Contextualized Machine Learning (ML), a paradigm for learning heterogeneous and context-dependent effects. 
Contextualized ML estimates heterogeneous functions by applying deep learning to the meta-relationship between contextual information and context-specific parametric models. 
This is a form of varying-coefficient modeling that unifies existing frameworks including cluster analysis and cohort modeling by introducing two reusable concepts: \emph{a context encoder} which translates sample context into model parameters, and \emph{sample-specific model} which operates on sample predictors. 
We review the process of developing contextualized models, nonparametric inference from contextualized models, and identifiability conditions of contextualized models. 
Finally, we present the open-source \texttt{PyTorch} package \package.
\end{abstract}

\section{Introduction}
\label{sec:introduction}
\begin{figure}[htb]
    \centering
    \includegraphics[width=0.8\textwidth]{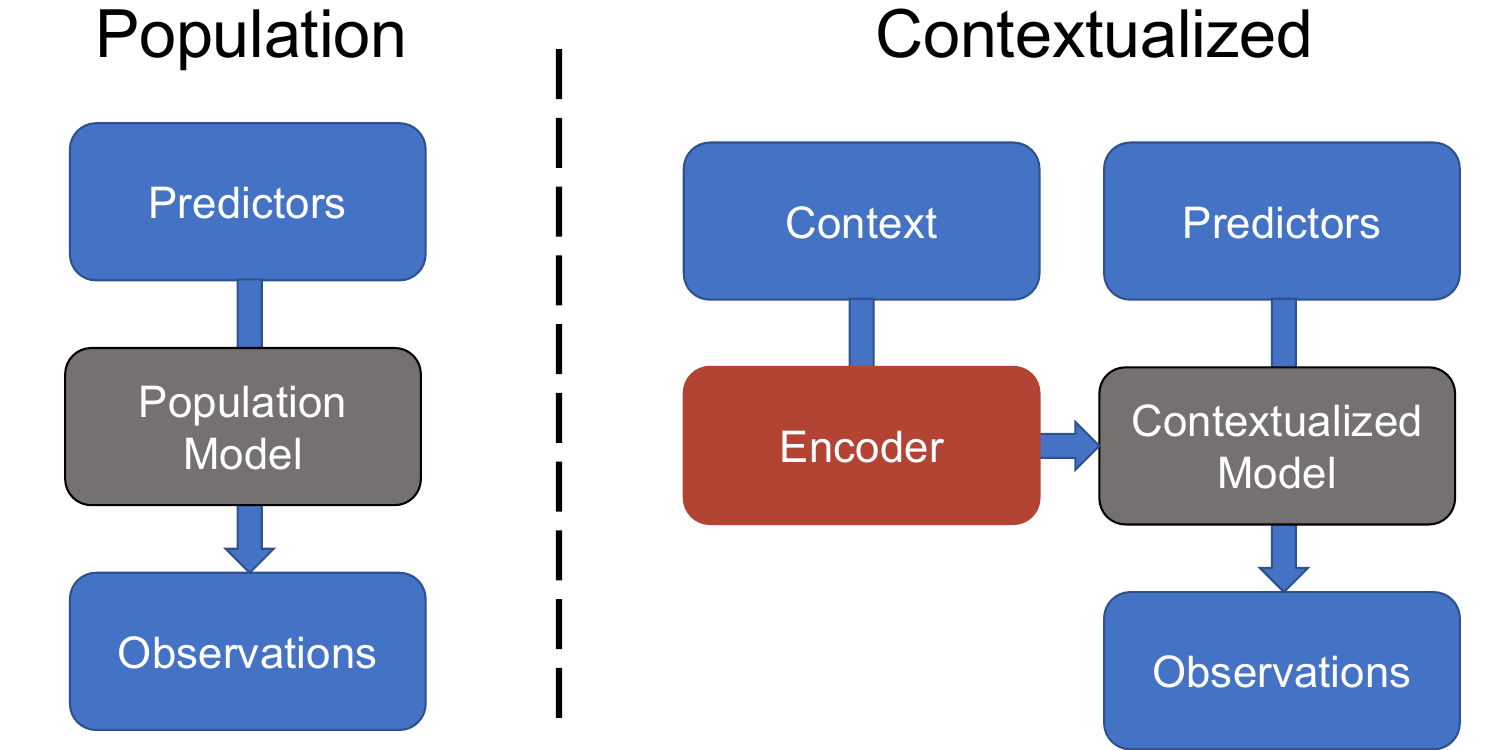}
    \caption{Contextualized paradigm. 
    Rather than using a single population model which operates identically for all contexts, Contextualized ML estimates a locally-optimal model for each context. 
    The heterogeneity of the population is captured by an encoder that translates sample context into sample-specific parameter values. 
    The contextualized model has the advantage of sharing information across the population while reducing model bias.}
    \label{fig:main_idea}
\end{figure}

Contextualized ML (Figure~\ref{fig:main_idea}) aims to learn the meta-effects of contextual information on parametric context-specific models by estimating a context encoder which translates sample context into sample-specific models.
By embracing the heterogeneity and context-dependence of natural phenomena, contextualized ML provides representational capacity while retaining the glass-box nature of statistical modeling. 
Contextualized models can be learned by simple end-to-end backpropagation because they are composed of differentiable building blocks. 
In the following, we study this paradigm, analyze identifiability and nonparametric inference through Contextualized ML, and provide a Python toolkit.

\subsection{Motivation}
\label{sec:motivation}
Modern applications of artificial intelligence are often characterized by training unconstrained ML models on large datasets. 
These datasets are composed of overlapping groups of samples, either explicitly (e.g. the large dataset is created by combining multiple datasets) or implicitly (e.g. the samples belong to latent sub-populations). 
To be generalizable, population models tend to prefer global patterns over localized effects, a problem when localized effects are critical to understanding complex processes such as in applications to computational biology (e.g. samples comprise latent cell types) and precision medicine (e.g. patients comprise latent disease subtypes). 
When faced with a localized effect, population-level models can either ignore the effect or encode the localized effect as an interaction of input variables. 
Neither of these solutions are attractive: ignoring the effect is high-bias while fitting unrestricted interactions is high-variance. 

Thus, we propose to use meta-models to generate context-specific parametric models. In this way, we can reason about the context-specific parameters and summarize meta-phenomena as explicit meta-models. 
This strategy often allows one to tackle ML challenges with more interpretable models, rather than trying to improve results by gathering more data or by opting for more complex models.

\paragraph{Towards Precision Medicine}
Precision medicine seeks to understand the patterns of differentiation between patients such that appropriate care can be provided for each individual. 
However, cohort-level models estimate the same effects for all patients in a cohort, ignoring sub-cohort heterogeneity.  
Since patients have different histories, environments, and disease sub-types, cohort-level models cannot appropriately model the patient journeys. 
As \cite{10.1145/3290605.3300468} found in clinical evaluation of a predictive model: 
``Some [doctors] voiced strong concerns that using [a ML model] was the same as applying `populational statistics' to individual patient decision making. They felt this was unethical.'' 
Thus, we seek to estimate models which adapt to patient context and drive personalized understanding. 
By estimating model parameters as functions of sample context, we can make principled sample-specific inferences. 

\paragraph{Towards Intelligible Artificial Intelligence}
Some applications of AI are limited due to strict requirements for intelligible and transparent decisions. 
Large population-level models with many implicit interaction effects can be difficult to interpret, and while post-hoc procedures to approximate the large model with locally-interpretable models \cite{ribeiro2016should} can provide approximations of the model, such approximations do not guarantee capturing the exact behavior of the population model. 
We propose to approach the same endpoint more directly: by learning contextualized models from the beginning, we achieve direct interpretability without requiring post-hoc interpretation of a black-box model.

\paragraph{Motivating Example}
Let us review the motivating example of \cite{lengerich2019learning}: understanding election outcomes at the local level. 
Given candidate representations, we wish to predict and understand the factors driving the candidate's vote proportion in a particular locality (e.g. county, township, district, etc.). 
One approach would be to partition the dataset into similar localities and then estimate cohort-specific models for each partition. 
Unfortunately, by building independent models for each county, we would fail to share information between related counties, forcing us to pool together some localities with fewer samples even though they may have distinct characteristics. 
This simultaneous loss of power and predictive accuracy is typical of modeling large, heterogeneous datasets with homogeneous models. 

Alternatively, instead of seeing these localities as discrete groups, we may embrace the data heterogeneity by modeling the $i$th county using a regression model $f(X_i;\Phi(C_i))$, where $\Phi(\cdot)$ is a parameter-generating function. 
This \emph{contextualized} modeling allows us to train accurate models using only a single sample from each county---this is useful in settings where collecting more data may be expensive (e.g. biology and medicine) or impossible (e.g. elections and marketing). 
By allowing the context to be sample-specific, $f$ no longer needs to be complex, and simple linear and logistic regression models will suffice, providing useful and interpretable models for each sample.

\section{Contextualized Machine Learning}

Contextualized ML estimates heterogeneous effects as distributions that adapt to context:
\begin{align}
    Y|X \sim \mathbb{P}_{\Phi(C)}.
\end{align}
That is, contextual data $C$ is transformed into a conditional distribution by a learnable function $\Phi$. 
For example, in this notation, the linear varying-coefficient model $Y|X \sim \text{N}(X\beta C^T, \sigma^2)$ \cite{hastie1993varying}  becomes the contextualized regression model:
$$\Phi(C) := \beta C^T, \qquad \mathbb{P}_{\Phi} = \text{N}(X\Phi(C), \sigma^2), $$
with $\beta \in \R^{p \times m}$ transforms context $C^T \in \R^{m \times 1}$ into sample-specific parameters. 
The free parameters are $\beta$ and $\sigma^2$, values of which can be estimated by backpropagation. 
This can be extended to accommodate heteroskedastic noise, e.g. by modeling noise as a separate function of context
$$\Phi(C) := (\beta C^T, \phi C^T), \qquad P_{\Phi} = \text{N}(X\Phi(C)_1, \Phi(C)_2),$$ 
or uncertainty in the sample-specific parameters, e.g. by a simple mixture
$$\Phi(C) := (\beta_1 C^T, \ldots, \beta_m C^T), \qquad P_{\Phi} = \sum_{i=1}^m \text{N}(X\Phi(C)_i, \sigma^2).$$
In this way, Contextualized ML amplifies the varying-coefficient paradigm by applying the power of deep learning and auto-differentiation.

\subsection{Contextualizing Models}
\begin{figure}[htb]
    \centering
    \includegraphics[width=0.95\textwidth]{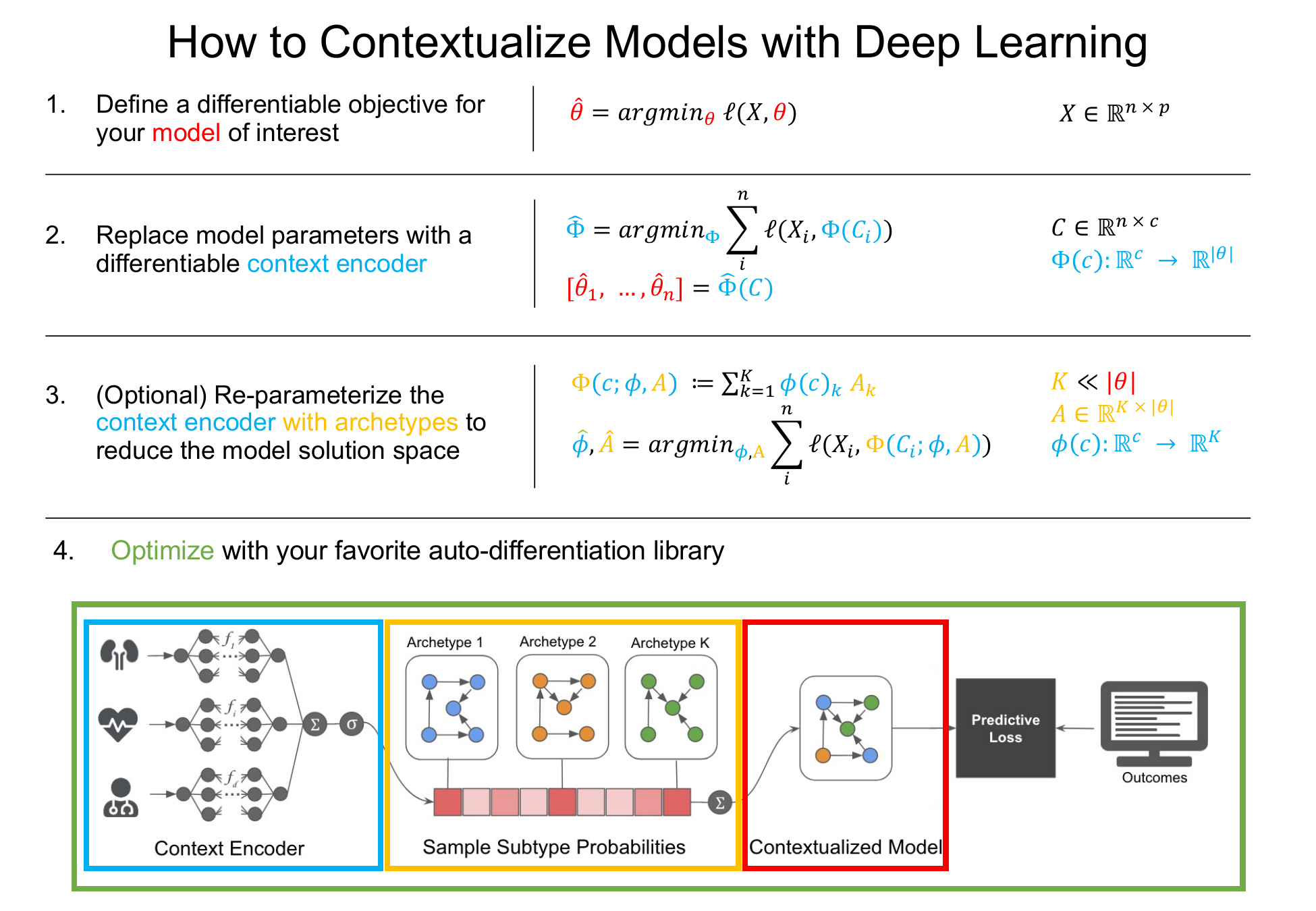}
    \caption{How to contextualize a cohort-based estimator. \textbf{(1)} Define a differentiable objective function for each sample-specific model (red). \textbf{(2)} Define a differentiable context encoder to generate sample-specific parameters (blue). \textbf{(3)} Re-parameterize the context encoder to reduce solution space (yellow). \textbf{(4)} Optimize end-to-end (green). 
    \label{fig:how_to_contextualize}
    }
\end{figure}

The general approach to designing contextualized versions of cohort-based estimators is summarized in Figure~\ref{fig:how_to_contextualize}. 
There are two potentially difficult steps in this process: defining a differentiable objective function, and designing a context encoder which operates on a tractable model solution space. 
While differentiable objective functions are problem-specific, there are a few general tricks which can often be used to improve the learnability of the deep context encoder $\Phi(C)$.

\paragraph{Restrictive Context Encoders}
Using a smaller class of context encoders can improve estimation. 
In practice, surprisingly simple forms of models can often be effectively used. 
For example, neural additive models \cite{agarwal2021neural}, which are differentiable forms of additive models, can be used as context encoders to eliminate interaction effects between contextual features and enable feature-specific interpretability of context--parameter links.

\paragraph{Archetype-Based Modeling}
Archetype-based modeling can reduce the dimensionality of the output of $\Phi$. 
By representing sample-specific models as weightings of $k$ archetypes the context encoder only needs to output a vector of size $k$, rather than the full model parameterization.
$$ \Phi(C) := \sum_{k=1}^K \phi(C)_k A_k$$
Furthermore, by restricting the archetype weightings to be non-negative and sum to 1 (e.g. by applying a softmax to $\phi$), the sample-specific models are a convex combination of the archetypes and can be interpreted as subtype probabilities with archetypes corresponding to subtype extrema. 

\paragraph{Regularizing Toward Population Models}
By simultaneously modeling all contexts, contextualized models can be encouraged to stay closer to the population model. 
Let $t \in T$ index a task-specific distribution $\mathbb{P}_t(Y|X)$. 
Multitask learning \cite{caruana1997multitask,breiman1997predicting} seeks to improve the estimation of each $\mathbb{P}_t(Y|X)$ by sharing power between distinct tasks $t$. 
Theorem 2 of \cite{lengerich2020sample} shows that the task-specific distribution is the sum of the overall distribution and a task-specific pure interaction effect:
\begin{equation}
    Y|X, t = Y|X + \rho(Y | X, t)
\end{equation}
where $\rho(Y | X, t)$ is a pure interaction effect \cite{lengerich2020purifying}. 
Thus, as sample context is an implicit task specifier, we see that context-specific estimators $Y|X, C$ also provide estimates of $Y|X$ and $\rho(Y|X, C)$ and so by regularizing against interactions between $X$ and $C$ (e.g. with Dropout \cite{lengerich2022dropout}), we can encourage similarity in the context-specific distributions and encourage them to be closer to the population model. 
Finally, we can use the purification \cite{lengerich2020purifying} to recover the task-specific interaction and the main effects from $Y|X, C$.

\newcommand{\param}{\theta}

\subsection{Related Work}
One of the earliest ways to model sample-specific parameters as the output of a learnable function was the linear varying-coefficients (VC) model~\citep{hastie1993varying} in which regression parameters are produced by a linear function of covariate values, e.g. 
$f(x; z) = \ip{x, \theta z}$, with $\theta \in \mathbb{R}^{P \times K}$ for $x \in \mathbb{R}^{P}$, $z \in \mathbb{R}^{K}$.
In short, contextualized ML combines the adaptability of the VC model with the power of modern ML architectures by using deep neural network as context-encoders. 
This combined approach was first proposed to improve interpretability of deep learning models \citep{al2017contextual} and has achieved good performance on varied tasks including survival prediction \citep{al2018personalized} and language modeling \citep{platanios2018contextual}. 
There have also been attempts to provide a nonparametric parameter-generating function by distance-matching regularization \cite{lengerich2018personalized,lengerich2019learning} which proposes that there is a distance metric on contextual information which approximates Euclidean distance between sample parameters (i.e. $\norm{\theta_i - \theta_j} \approx d_C(C_i, C_j)$ for samples $i,j$). 
While this regularization-based scheme provides extra flexibility by obviating the requirement of a parameteric context encoder, it also precludes end-to-end training due to the lack of a differentiable context encoder. 

\subsubsection{Alternative Approaches}
\paragraph{Sample-Specific Models as Deviations}
Recent work has also developed sample-specific estimators as independent deviations from a population model~\citep{pmlr-v72-jabbari18a,kuijjer2019estimating,liu2016personalized,li2018learning}.
This is particularly useful for structured models in which prior knowledge of the graph structure can enable efficient testing of sample-specific deviations. 
However, estimating sample-specific models as deviations requires $\mathcal{O}(n)$ estimation procedures for $n$ samples and does not share power between the estimators. 
As a result, these approaches are more applicable to domains with fewer samples and less informative contextual data.

\paragraph{Heterogeneous Samples}
Statistical tests~\citep{gu2018testing,liu2003testing,charnigo2004testing} can identify whether a cohort contains heterogeneous samples, enabling the identification of partitions that induce accurate group-based models. 
This perspective is well-suited to situations with a small numbers of groups or pre-defined partitions. 
However, if there are many groups relative to the number of samples, group-based modeling becomes high-variance and if samples arise from a continuous combination rather than discrete partitions, group-based models become high-bias. 
In such situations, higher resolution via multitask learning is required.

\paragraph{Post-Hoc Model Interpretations}
Methods of post-hoc interpretation often seek to explain complex models by estimating local approximations \citep{hendricks2016generating,shrikumar2016not,ribeiro2016model,lakkaraju2017interpretable}. 
For example, Local-Interpretable Model-Agnostic Explanation (LIME) \citep{ribeiro2016should} constructs local interpretations for each sample by training a linear model to approximate the outputs of a black-box model in a particular neighborhood. 
These local models are interpretable and approximate the output of any model, but are constrained to explain only a fixed black-box population model.
In contrast, contextualized regression directly estimates local models, enabling dynamic collections of models that retain local interpretability.

\subsection{Benefits of Contextualized ML}
In the following, we demonstrate a few benefits of contextualized ML. 
More details and reproducible demos for these perspectives are available in this \href{https://contextualized.ml/docs/demos/benefits.html}{Jupyter notebook}.

\begin{figure}[htb]
    \centering
    \includegraphics[width=0.7\textwidth]{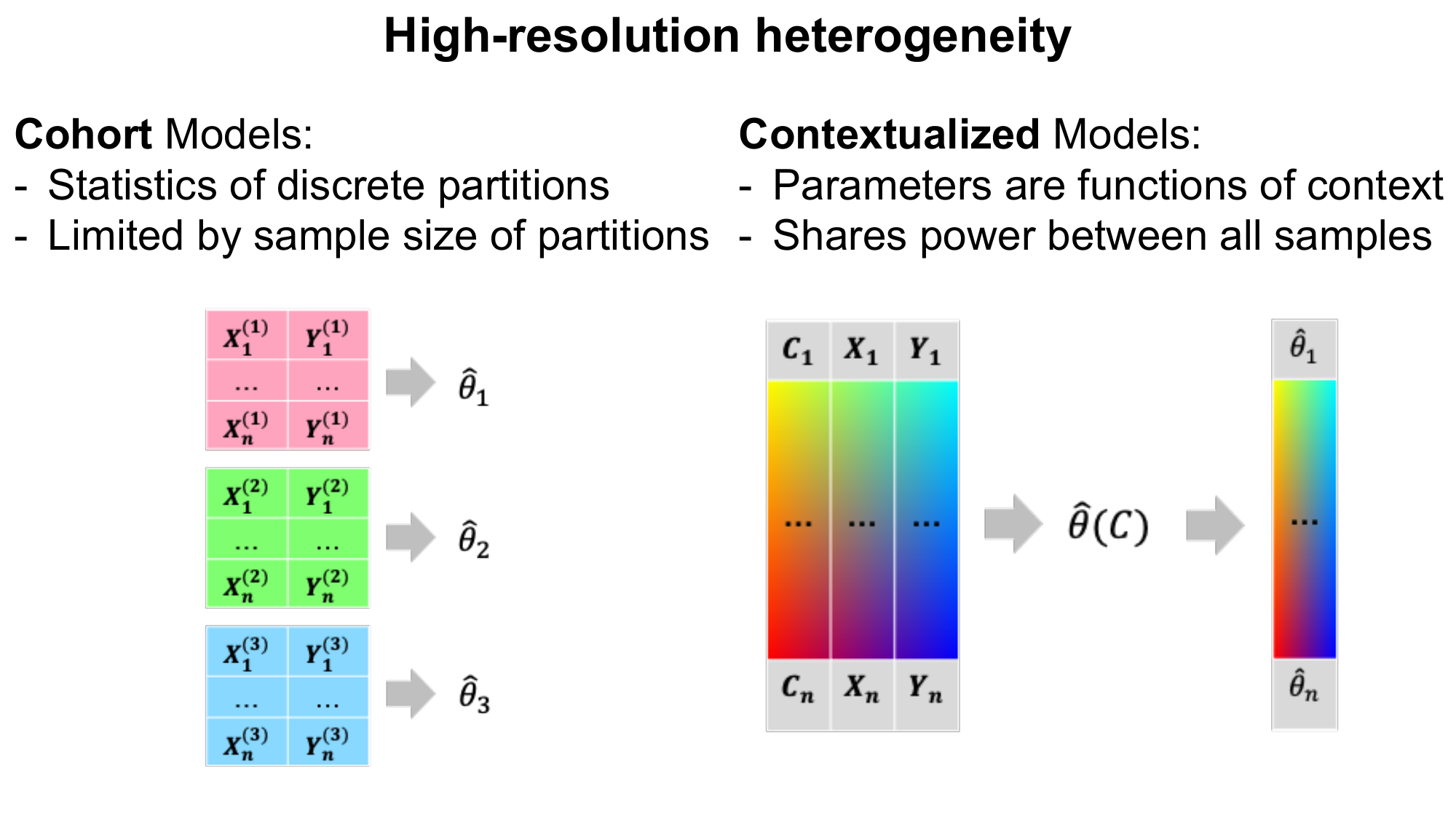}
    \caption{By sharing power between samples, contextualized ML recovers heterogeneous effects at resolutions which are finer-grained than can be done by partition-based cohort models.}
    \label{fig:high_resolution}
\end{figure}

\paragraph{Contextualized ML Enables High-Resolution Heterogeneity}
By sharing information between all contexts, contextualized learning is able to estimate heterogeneity at fine-grained resolution (Figure~\ref{fig:high_resolution}). 
Cluster or cohort-based models treat every partition independently, limiting heterogeneity to coarse-grained resolution where there are large enough cohorts for independent estimation. 
For example, this ability was exploited for context-specific Bayesian networks \cite{lengerich2021notmad} (while cohorts models would require $\mathcal{O}(p^2)$ samples in each cohort) to reconstruct patient-specific gene expression networks. 

\begin{figure}[htb]
    \centering
    \includegraphics[width=0.8\textwidth]{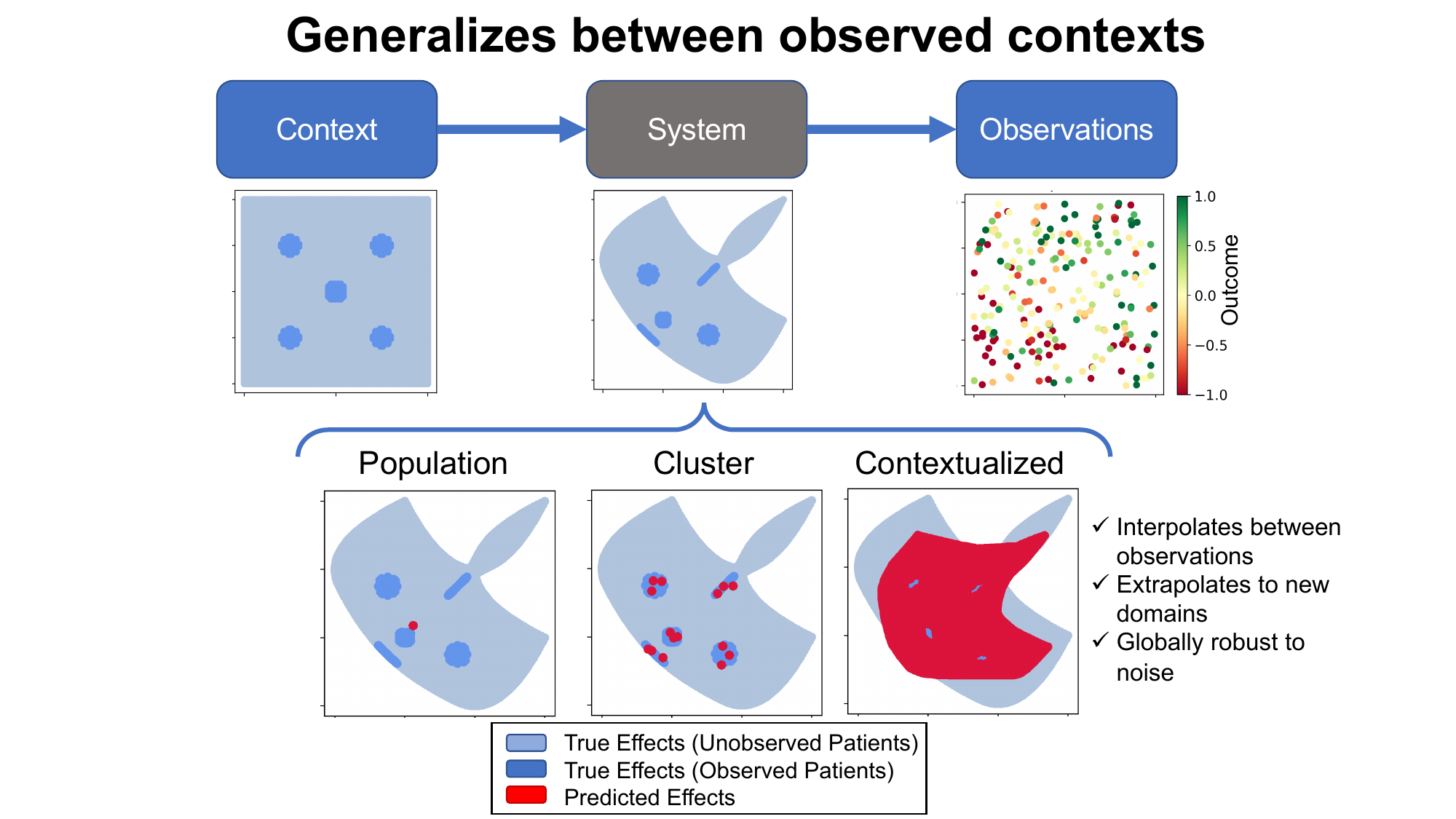}
    \caption{By learning the meta-relationship between context and model parameters, contextualized ML enables interpolation between observed contexts.}
    \label{fig:extrapolates}
\end{figure}

\paragraph{Contextualized ML Interpolates Between Observed Contexts}
By learning to translate contextual information into model parameters, contextualized models learn about the meta-distribution of contexts (Figure~\ref{fig:extrapolates}). 
As a result, contextualized models can adapt to contexts which were never observed in the training data, either interpolating between observed contexts or extrapolating to new contexts for which the meta-relationship between context and local parameters holds the same as in the training data. 

\begin{figure}[htb]
    \centering
    \includegraphics[width=0.7\textwidth]{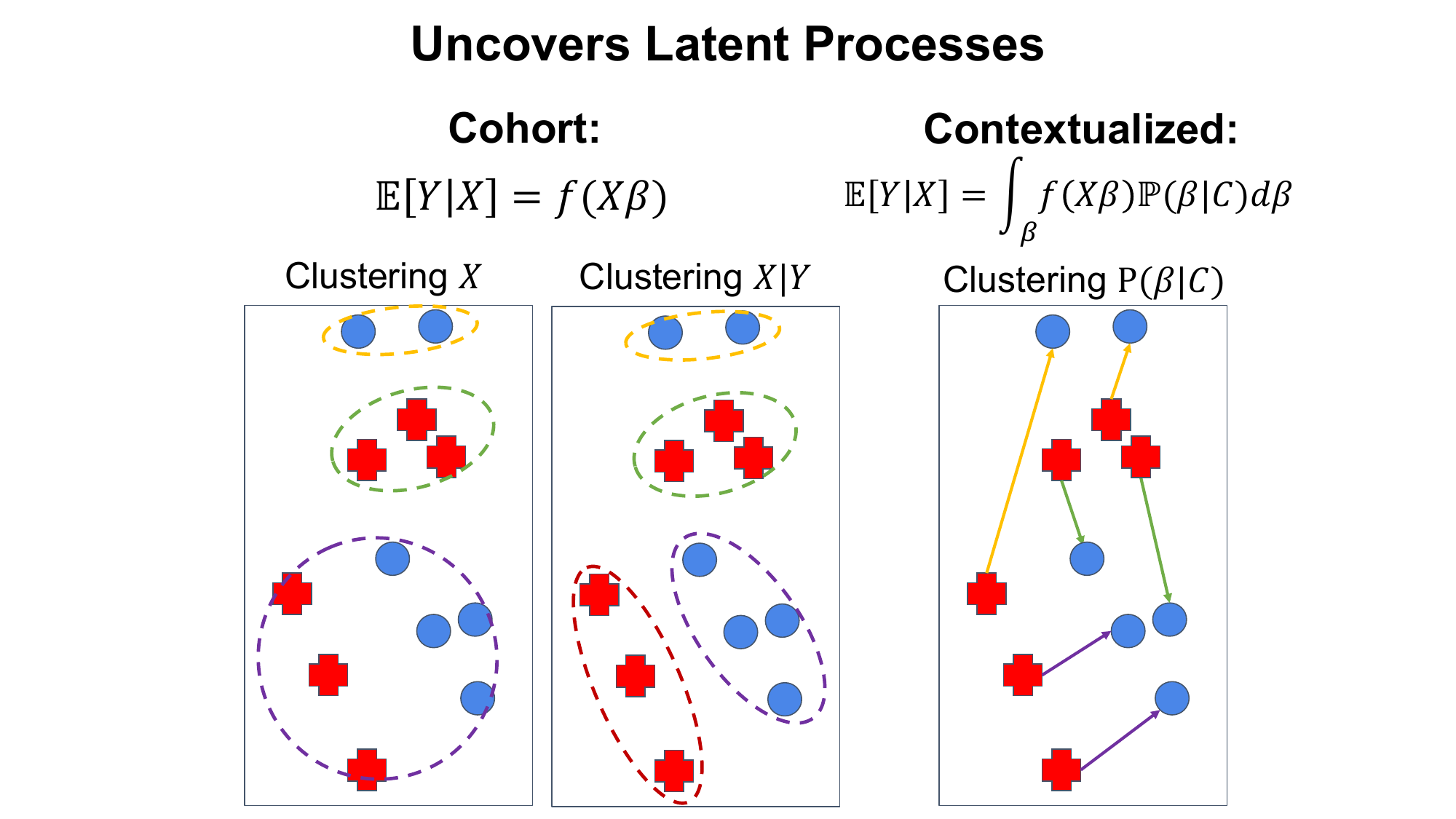}
    \caption{By estimating a contextualized model for each sample, contextualized ML uncovers important factors and latent processes in heterogeneous populations.}
    \label{fig:latent_processes}
\end{figure}

\paragraph{Contextualized ML Enables Analysis of Latent Processes}
Cluster or cohort models which are inferred by partitioning samples into groups make assumptions of IID data within each group. 
This approach works well when contexts are discrete, low-dimensional, and every context-specific population is well observed, but in many complex processes, contexts are continuous, high dimensional, and sparsely observed.
When cluster or cohort approaches are applied in these circumstances, downstream modeling tasks are distorted by mis-specification, where many non-IID samples are funneled into a single model.
Consequently, theoretical guarantees about how well a cluster or cohort model can represent IID populations often do not apply in light of real-world heterogeneity.
In contrast, contextualized learning provides a way to estimate latent, non-IID models for all samples with minimal assumptions about the grouping or clustering of these samples (Figure~\ref{fig:latent_processes}). 
Samples can then be grouped on the basis of model parameters and distributional differences to produce clusters in the latent model space underlying each sample. 
Contextualized ML intuitively recovers latent structures underlying data generation in a way \textit{a priori} clustering cannot.
Allowing downstream models to determine the grouping of samples rather than upstream contexts replaces traditional cluster analysis with contextualized analysis clusters.

\subsection{Python Package}
Contextualized GLMs are implemented in \package~with easy interfaces. These GLMs take the form
\begin{align}
    \mathbb{E}[Y|X, C] = f\left(X\Phi(C)\right), 
\end{align}
where $\Phi(C)$ is a deep context encoder, 
For example, contextualized linear regression: 
\begin{align}
    \mathbb{E}[Y|X,C] = X\Phi(C),
\end{align}
is available by the \texttt{ContextualizedRegressor} class:
\begin{lstlisting}
    from contextualized.easy import ContextualizedRegressor
    model = ContextualizedRegressor()
    model.fit(C_train, X_train, Y_train)
\end{lstlisting}
Similarly, contextualized logistic regression:
\begin{align}
    \mathbb{E}[\text{Pr}(Y=1)|X,C] = \sigma(X\Phi(C))
\end{align}
is available by the \texttt{ContextualizedClassifier} class: 
\begin{lstlisting}
    from contextualized.easy import ContextualizedClassifier
    model = ContextualizedClassifier()
    model.fit(C_train, X_train, Y_train)
\end{lstlisting}

Common constructor keywords include:
\begin{itemize}
    \item \texttt{n\_bootstraps}: Number of bootstrap resampling trajectories to use.
    \item \texttt{encoder\_type}: \texttt{mlp}, \texttt{ngam}, or \texttt{linear}, which type of model to make as context encoder. Alternatively, users may pass in their own encoder as a \texttt{PyTorch} module.
    \item \texttt{loss\_fn}: A function to calculate loss.
    \item \texttt{alpha}: non-negative float, regularization strength.
    \item \texttt{mu\_ratio}: float in range (0.0, 1.0), governs how much the regularization applies to context-specific parameters or context-specific offsets.
    \item \texttt{l1\_ratio}: float in range (0.0, 1.0), governs how much the regularization penalizes $\ell_1$ vs $\ell_2$ parameter norms.
\end{itemize}

Common fitting keywords include:
\begin{itemize}
    \item \texttt{max\_epochs}: positive number, the maximum number of epochs to fit. Early stopping is turned on by default.
    \item \texttt{learning\_rate}: positive float, default is 1e-3.
    \item \texttt{val\_split}: float in range (0.0, 1.0), how much of the data to use for validation (early stopping).
\end{itemize}

\section{Nonparametric Inference from Contextualized Models}
Contextualized ML provides a framework to estimate nonparametric densities by viewing the composite densities as combinations of local parametric distributions. 
Let us consider a regression $Y|X \sim p(f(X))$. 
This regression may be considered nonparametric in two non-exclusive respects:
\begin{itemize}
    \item the transmission function $f$ may not be well-represented by a parametric family, or
    \item the distribution $p$ may not be well-represented by a parametric family. 
\end{itemize} 
Contextualized linear models can be used to recover either of these forms of nonparametric models.

\begin{figure*}[htb]
\centering
\includegraphics[width=0.8\textwidth]{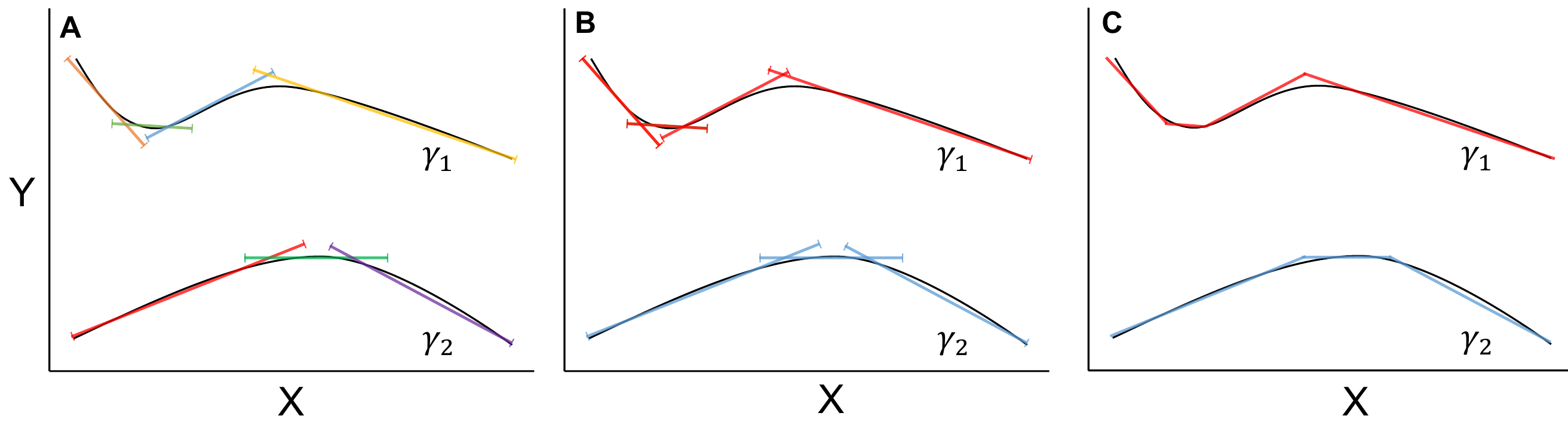}
\caption{\textbf{(A)} We observe data arising from a mixture of nonparametric context-specific densities $\gamma_1$ and $\gamma_2$, then  fit context-specific regression functions. \textbf{(B)} These context-specific atoms can be clustered and \textbf{(C)} smoothed into component to produce a nonparametric mixture model. The clustering recovers $\gamma_1$, $\gamma_2$ if the components are well-separated. \label{fig:np_transmission}}
\end{figure*}

\paragraph{Contextualized Linear Models Represent Nonparametric Transmission Functions}
First, contextualized linear models can represent nonparametric transmission functions by allowing coefficients to vary with context. 
As $\Phi(C) = \E_{X|C}[\frac{\partial\E[Y|X, C]}{\partial X}]$, we can view $\Phi$ as a differential expression describing $\frac{\partial\E[Y|X, C]}{\partial X}$ and reconstruct smooth, differentiable transmission functions by stitching together context-specific linear transmission functions (Figure~\ref{fig:np_transmission}). 
This approach approximates a conditional mixing distribution $\Gamma(x) =\sum_{k=1}^K\lambda_k(x)\gamma_k(x)$ of $K$ true mixtures by fitting an overfitted mixture of $L \gg K$ atoms and then clustering these $L$ atoms into $K$ groups such that each group approximates $\gamma_k$. 
Based on this clustering, we can define a new mixing measure whose atoms are close to some $\gamma_k$ for each $k$. 
This mixing measure will converge to $\Gamma$ as $L \rightarrow \infty$, allowing us to approximate $\Gamma$ to arbitrary precision.
This framework is illustrated in Figure~\ref{fig:np_transmission} and retains identifiability of the nonparametric transmission functions under reasonable assumptions of component separation \cite{aragam2018identifiability}. 

\begin{figure}[htb]
    \centering
    \includegraphics[width=0.8\textwidth]{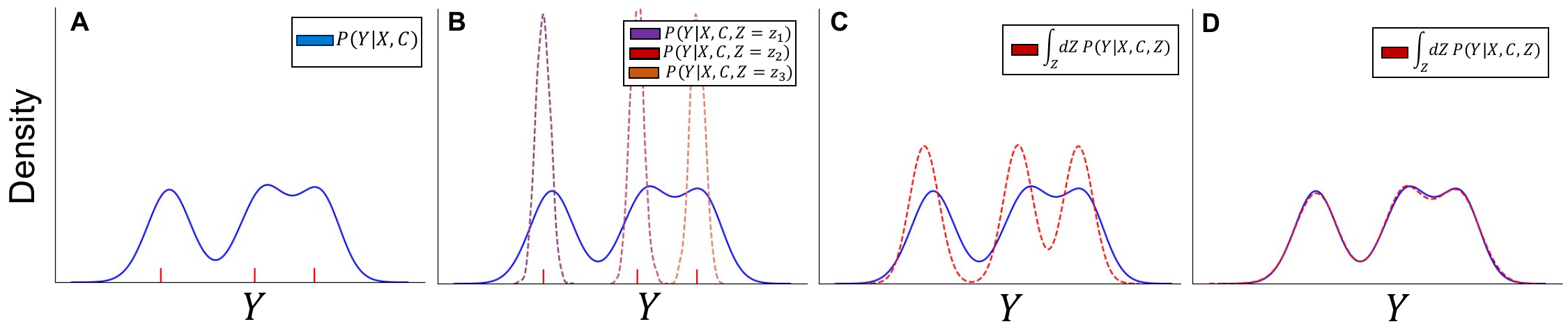}
    \caption{Pseudo-sampling procedure for representing nonparametric distributions. {\textbf{(A)}} The density $Y|X,C$ may not be well-approximated by a Gaussian distribution. \textbf{(B)} To overcome this, we can extend context by introducing noise variable $Z$ to psuedo-sample localized overfitted distributions centered at each sample observation (red vertical tick marks along the horizontal axis). {\textbf{(C)}} By integrating over the introduced noise variable, we approximate the nonparametric distribution. \textbf{(D)} Approximation improves as the number of observations increases.}
    \label{fig:nonparametric}
\end{figure}

\paragraph{Contextualized Models Represent Non-Gaussian Outcomes}
Second, contextualized models can represent non-Gaussian outcomes by summing context-specific Gaussian distributions. 
As locally-Gaussian distributions are universal approximators \citep{aragam2018identifiability}, any outcome distribution can be constructed by combining context-specific Gaussian distributions. 
If $Y|X,C$ is not well-approximated as a Gaussian distribution, 
we can \emph{pseudo-sample} extra noise variables $Z$ which localize the distribution such that $Y|X, C, Z$ is well-approximated as a Gaussian (Figure~\ref{fig:nonparametric}). 
In an extreme case, each value of $Z$ can identify an individual training sample with corresponding locally-Gaussian outcome distributions that sum to form a meaningful composite distribution. 
As with many latent variable problems, in test samples we cannot identify which value of $Z$ would be most correct; by integrating over all pseudo-sampled values of $Z$ we can reconstruct the nonparametric uncertainty. 

\section{Identifiability of Contextualized Models}
\label{sec:identifiability}

When seeking to understand contextualized models, we are interested in questions of identifiablity: how many sets of sample-specific models could equivalently recapitulate the observed data? 
For example, we know that both population and group-level linear models are identifiable in common conditions \cite{reiersol1950identifiability,ljung1994global,hennig2000identifiablity}, but sample-specific models without covariates or constraints are not identifiable. 
Does the process of generating contextualized models from a shared context encoder induce identifiability? 
Here, we present an informal, graphical view of identifiability of contextualized models that suggests that identifiability is influenced by the flexibility of \emph{both} the context encoder and the sample-specific models.


\paragraph{Notation}
Let us consider a sample-specific model class parameterized by $\theta \in \mathcal{H} \subset \R^p$ which induces 
solution space $s(x,y) = \{\theta \in \mathcal{H}: h(x; \theta) = y\}$ for sample $x,y$. 
A dataset $\mathcal{D} = C,X,Y = [(C_1,X_1, Y_1), \ldots, (C_n,X_n,Y_n)]$ induces a list of solution spaces $S(\mathcal{D}) = [s(X_1,Y_1),\ldots,s(X_n,Y_n)]$. 
For context encoders parameterized by $\phi \in \mathcal{G} \subset R^m$, let $G(\mathcal{D}) = \{\phi \in \mathcal{G}: g(C_i; \phi) \in s(X_i, Y_i)~\forall~i~\in [1,\ldots,n] \}$ be the set of allowable context encoders for this dataset. 
When $|G(\mathcal{D})| \leq 1$, there is at most one context encoder which maps each sample's context observation to its corresponding solution space, and we can say that the contextualized models are identifiable for this dataset.

\begin{figure}[htb]
    \centering
    \includegraphics[width=\textwidth]{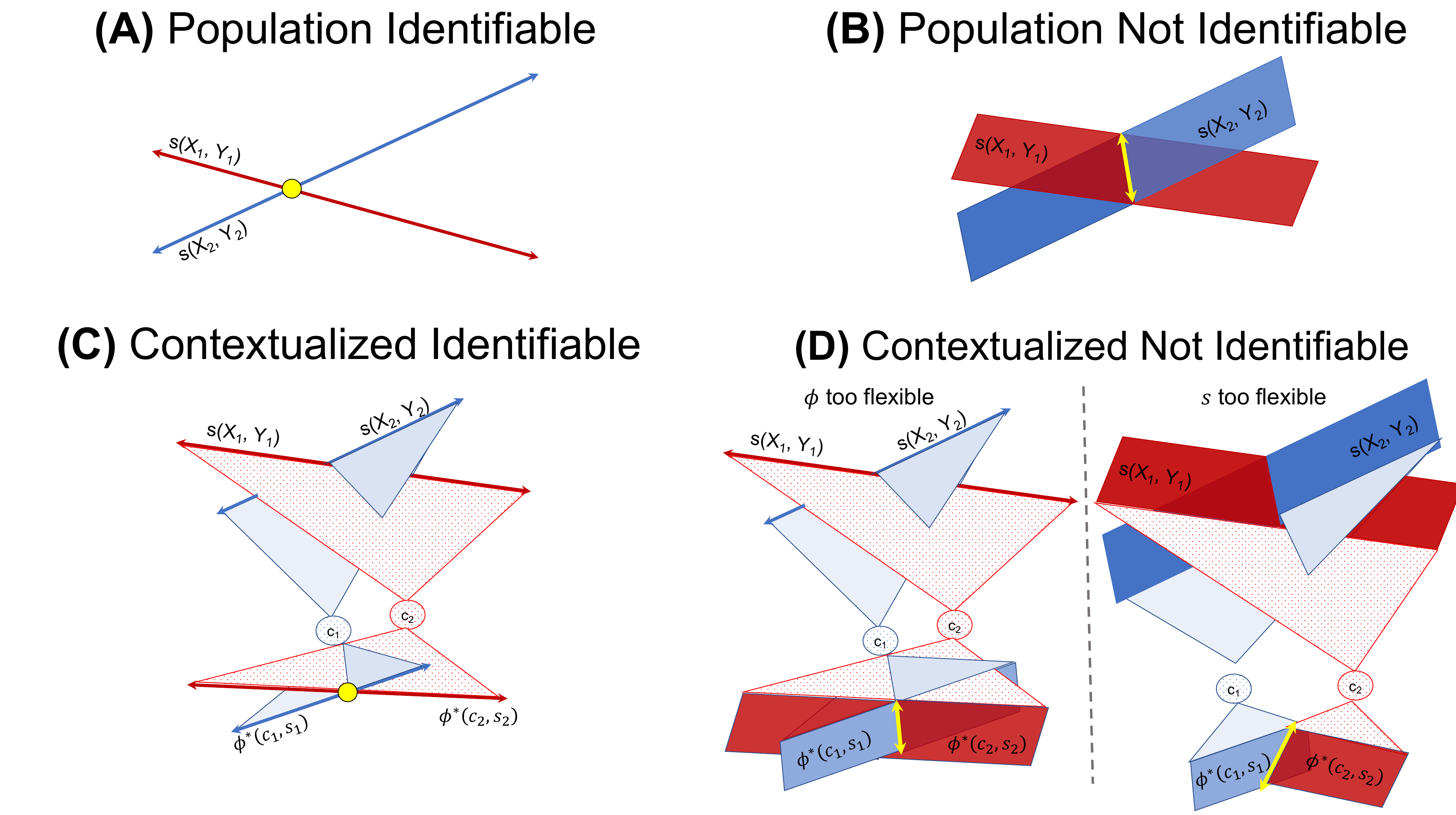}
    \caption{Graphical depiction of identifiablity. \textbf{(A-B)} Population models are defined by the intersection of sample-specific solution spaces. In each pane, we have two solution spaces $s(X_1,Y_1)$ and $s(X_2,Y_2)$ with their intersection marked in yellow. If each sample-specific solution space has $p$ dimensions of freedom, then identifiability requires $p$ intersecting solution spaces. \textbf{(C-D)} Contextualized models are defined by the intersection of the allowable context encoder spaces, and hence can lose identifiability for either of two reasons: excess flexibility in the context encoder $\phi$ or excess flexbility in the sample-specific solution spaces $s$.}
    \label{fig:contextualized_identifiability}
\end{figure}

\paragraph{Identifiability of Population Models} As a comparison, let us first consider population models from this perspective. 
Population models, which share $\theta$ for all samples, can be seen as constant context encoders: $g(c; \phi) = \phi$. 
Thus, the set of allowable context encoders for a dataset is $G_{\text{pop}}(\mathcal{D}) = \{\theta \in \mathcal{H}: h(X_i; \theta) = Y_i~\forall~i~\in [1,\ldots,n] \}= \bigcap_{i=1}^n s(X_i, Y_i)$, i.e. 
identifiability of a population model is defined by the size of the intersection of the sample-specific solution spaces. 
For example, identifiability of a linear regression model is determined by how many sample-specific solution spaces (hyperplanes) coincide and how many intersect: if $p$ solution spaces intersect, the linear model of $p$ variables is identifiable. 
For linear regression with $p=2$, $n\geq 2$ is sufficient to provide identifiability (Figure~\ref{fig:contextualized_identifiability}A). 
For linear regression with $p=3$, the sample-specific solution spaces have $2$ degrees of freedom and hence 2 samples can only constrain $G_{\text{pop}}(\mathcal{D})$ to a 1-dimensional subspace (Figure~\ref{fig:contextualized_identifiability}B). 

\paragraph{Identifiability of Contextualized Models}
For contextualized models, we are interested in the set of allowable context encoders for each sample: $\phi^*(c, s) = \{\phi \in \mathcal{G}: g(c; \phi) \in s\}$. 
The intersection of these sample-specific sets of allowable context encoders determines the allowable context encoders for the data: $G_{\text{contextualized}}(\mathcal{D}) = \bigcap_{i=1}^n\phi^*(C_i, s(X_i, Y_i))$. 
The dimension of $\phi^*(c,s)$ is upper-bounded by the product of the dimension of $s$ and a measure of the redundancy in the context encoder (how many ways can each solution be generated). 
Ignoring pathological collinearity, this suggests a simple heuristic for contextualized identifiability: $n > d_gd_s$, where $d_g$ is the degree of redundancy in the context encoder class and $d_s$ is the number of degrees of freedom in each solution space $s$. 



A few examples may make this heuristic more concrete. 
For population models, $d_g=1$ (only a constant function can return the same value for all inputs), and hence identifiability of population models are determined by the number of degrees of freedom in the solution space. 
For contextualized linear models, $d_s=p-1$, suggesting that $n> d_g(p-1)$ is a useful criterion for identifiability of contextualized linear models. 
For linear varying-coefficients models, this criterion becomes $n> m(p-1)$, which can be compared to traditional identifiability criteria for linear varying-coefficients models \cite{cai2000efficient,kuruwita2011generalized,yue2019identification,zhang2015estimation,hu2019estimation}.
With $m=1$ and $p=2$ (Figure~\ref{fig:contextualized_identifiability}C), 2 samples are sufficient for identifiability. 
Note that $m=1$ means that the context encoder operates on a single contextual variable; this single contextual variable is typically a vector of ones to accommodate offsets. 
For either $m=p=2$ (Figure~\ref{fig:contextualized_identifiability}D, left) and $m=1,p=3$ (Figure~\ref{fig:contextualized_identifiability}D, right), at least 3 samples are required for identifiability.

\section{Discussion}
We have examined Contextualized ML, a paradigm for context-specific inference of differentiable models. 
This framework provides a principled method for sample-specific inference and analysis of heterogeneous effects, and we have presented the package~\package~to make standard tasks of context-specific regression and context-specific network inference accessible to Python users.

Several research directions remain open. 
While deep learning-based context encoders and auto-differentiation libraries are useful to circumvent requirements of parametric assumptions and analytical solutions, there is no guarantee that this learning scheme is optimal. 
In addition, these methods rely on contextual data to accurately represent latent phenomena; extending methods to generate sample representations from more diverse data sources (e.g. foundation models) could improve the learned models. 
Beyond questions of estimation procedures, there are also open questions regarding the analysis of estimated sample-specific models. 
Once we have estimated sample-specific model parameters, what is the best way to summarize these new representations: should we cluster the estimated parameters, or is it best to present these models to users as sample-specific models? 
These questions scratch the surface of the wide potential that contextualized ML unlocks for improved methods of data analysis.

\subsection*{Acknowledgements}
We thank Wesley Lo, Jannik Deuschel, Juwayni Lucman, Alyssa Lee, and Aaron Alvarez for their contributions to the development and use of the Python package. We are also very grateful to Bryon Aragam, Maruan Al-Shedivat, Avinava Dubey, Amir Alavi, and Rich Caruana for valuable discussions.


\end{document}